\documentclass[letterpaper]{article} 
\usepackage{aaai24}  
\usepackage{times}  
\usepackage{helvet}  
\usepackage{courier}  
\usepackage[hyphens]{url}  
\usepackage{graphicx} 
\usepackage{natbib}  
\usepackage{caption} 
\usepackage{algorithm}
\usepackage{algorithmic}

\usepackage{amsmath}
\usepackage{amssymb}
\usepackage{color}
\usepackage{multirow}
\usepackage{pifont} 
\usepackage{subcaption}
\usepackage[table,xcdraw]{xcolor}
\usepackage[T1]{fontenc}
\usepackage{tabularx}

\usepackage{xcolor} 
\usepackage{algorithm}
\usepackage{algorithmic}
\usepackage{amsmath}
\usepackage{amssymb}
\usepackage{tabularx}
\usepackage{multirow}
\usepackage{rotating} 
\usepackage{caption}
\usepackage{subcaption}
\usepackage{pifont}
\usepackage{booktabs}
\usepackage{enumitem}
\usepackage{cleveref}
\usepackage{array}

\usepackage{newfloat}
\usepackage{listings}
\DeclareCaptionStyle{ruled}{labelfont=normalfont,labelsep=colon,strut=off} 
\floatstyle{ruled}
\newfloat{listing}{tb}{lst}{}
\floatname{listing}{Listing}
\title{Dynamic Feature Pruning and Consolidation for Occluded Person Re-Identification}
\author{
    Yuteng Ye\textsuperscript{1},
    Hang Zhou\textsuperscript{1},
    Jiale Cai\textsuperscript{1}, 
    Chenxing Gao\textsuperscript{1}, 
    Youjia Zhang\textsuperscript{1}, 
    Junle Wang\textsuperscript{2},
    Qiang Hu\textsuperscript{3},
    Junqing Yu\textsuperscript{1},
    Wei Yang\textsuperscript{1}\thanks{indicates corresponding author.}
}

\affiliations{
    \textsuperscript{1}Huazhong University of Science and Technology, Wuhan, China\\
    \textsuperscript{2}Tencent\\
    \textsuperscript{3}Shanghai Jiao Tong University, Shanghai, China\\
    \{yuteng\_ye, henrryzh, jaile\_cai, cxg, youjiazhang, yjqing, weiyangcs\}@hust.edu.cn; 
    wangjunle@gmail.com;
    qiang.hu@sjtu.edu.cn
}

\newcommand{\ie}{i.e., }
\newcommand{\eg}{e.g., }
\newcommand{\etal}{\textit{et al.}}

\usepackage{bibentry}

\begin{document}

\maketitle
\begin{abstract}
Occluded person re-identification (ReID) is a challenging problem due to contamination from occluders. Existing approaches address the issue with prior knowledge cues, such as human body key points and semantic segmentations, which easily fail in the presence of heavy occlusion and other humans as occluders.
In this paper, we propose a feature pruning and consolidation (FPC) framework to circumvent explicit human structure parsing. The framework mainly consists of a sparse encoder, a multi-view feature mathcing module, and a feature consolidation decoder.
Specifically, the sparse encoder drops less important image tokens, mostly related to background noise and occluders, solely based on correlation within the class token attention.
Subsequently, the matching stage relies on the preserved tokens produced by the sparse encoder to identify k-nearest neighbors in the gallery by measuring the image and patch-level combined similarity.
Finally, we use the feature consolidation module to compensate pruned features using identified neighbors for recovering essential information while disregarding disturbance from noise and occlusion.
Experimental results demonstrate the effectiveness of our proposed framework on occluded, partial, and holistic Re-ID datasets. In particular, our method outperforms state-of-the-art results by at least 8.6\% mAP and 6.0\% Rank-1 accuracy on the challenging Occluded-Duke dataset.

\end{abstract}

\section{Introduction}
\label{sec:intro}
Person Re-Identification (ReID) refers to the process of retrieving the same person from a gallery set under non-overlapping surveillance cameras~\cite{chen2017person, ye2021deep}, and has been making remarkable progress in tackling appearance change in deep learning era~\cite{wu2019deep, lavi2018survey}.
However, the re-identification of occluded persons remains a challenging problem because of two reasons: 1. the inference from wrongly included occluder features and 2. the partial or full absence of essential target features.
To tackle occlusions, many existing approaches explicitly recover human semantics, via human pose estimation~\cite{miao2019pose, gao2020pose, wang2022pose} or body segmentation~\cite{huang2020human}, as extra supervision to guide the network to focus on non-occluded features. Others~\cite{yu2021neighbourhood, xu2022learning} first partition input image into horizontal or vertical parts, and then identify the occlusion status of each part with off-the-shelf models (Mask-RCNN~\cite{he2017mask}, HR-Net~\cite{sun2019deep}), and finally recover occluded features from K-nearest neighbors in a gallery according to non-occluded features. Both strategies rely on extra modules for detecting occlusions and will easily fail in the presence of heavy occlusion and persons as occluders, and further background noise persists as the partition usually is very coarse.

Inspired by the recent advances in sparse encoders~\cite{rao2021dvit, liang2022evit}, we propose a feature pruning, matching, and consolidation (FPC) framework for Occluded Person Re-Identification which adaptively removes interference from occluders and background and consolidates the contaminated features.
%
%
Firstly, we send the query image into a modified transformer encoder with token sparsification to drop interference tokens (usually related to occluders and background) while preserving attentive tokens. 
Different from extra cue-based approaches that rely on prior information about human semantics, the sparse encoder exploits correlation properties on attention maps and generalizes better to various occlusion situations. In addition, our sparse encoder removes interference from the background as an extra benefit. Then, we rank the full tokens in the gallery memory according to their similarity with the query image. We obtain the gallery memory containing $[\mathrm{cls}]$ token and patch tokens via pre-training a vision transformer encoder. 
The similarity metric for matching is defined as the linear combination of image-level cosine distance and patch-level earth mover’s distance~\cite{rubner2000earth} for bridging the domain gap between the sparse query feature and holistic features in the gallery memory.
At last, we select the k-nearest neighbors for each query and construct multi-view features by concatenating averaged $[\mathrm{cls}]$ tokens and respective patch tokens of both the query and its selected neighbors. 
We send the multi-view feature into a transformer decoder for feature consolidation. We compensate the occluded query feature with gallery neighbors and achieve better performance.
During training, our method utilizes the entire training set as the gallery, a common practice in ReID literature~\cite{xu2022learning, yu2021neighbourhood}. While for inference, only the test set query and gallery images are used without any pre-filtering steps.
%
Our FPC framework achieves state-of-the-art performance on both occluded person, partial person, and holistic person ReID datasets. In particular, FPC outperforms the state-of-the-art by 8.6\% mAP and 6.0\% Rank-1 accuracy on the challenging Occluded-Duke dataset.

Our main contributions are as follows:

\makeatletter

\def\@listi{%

  \leftmargin\leftmargini

  \parsep 0pt%

  \topsep \parsep

  \itemsep \parsep}

\let\@listI\@listi

\@listi

\def\@listii{%

  \leftmargin\leftmarginii

  \labelwidth\leftmarginii

  \advance\labelwidth-\labelsep

  \parsep 0pt%

  \topsep \parsep

  \itemsep \parsep}

\def\@listiii{%

  \leftmargin\leftmarginiii

  \labelwidth\leftmarginiii

  \advance\labelwidth-\labelsep

  \parsep 0pt%

  \topsep \parsep

  \itemsep \parsep

  \partopsep \p@ \@plus\z@ \@minus\p@}

\makeatother
\begin{itemize}
		\item[$\bullet$] We introduce the token sparsification mechanism to the occluded person ReID problem, the first to the best of our knowledge, which avoids explicit use of human semantics and better prunes unrelated features.  
		\item[$\bullet$] We propose feature matching and consolidation modules to recover occluded query features from multi-view gallery neighbors. Compared to feature division approaches~\cite{yu2021neighbourhood, xu2022learning}, our module uses transformer tokens that naturally preserve connectivity and richness of the features.
		\item[$\bullet$]  We design a novel token based metric to measure the similarity of images by linearly combining the patch-level distance and the image-level distance.




\end{itemize}

\section{Related Work}
Our method is closely related to occluded person ReID and feature pruning in vision transformers.

\textbf{Occluded Person ReID}. 
The task of Occluded Person ReID is to find the same person under different cameras while the target pedestrian is obscured. The current methods are mainly divided into two categories, \ie extra-clues based methods and feature reconstruction based methods. 
The extra-clues based methods locate non-occluded areas of the human body by prior knowledge cues, \eg human pose~\cite{miao2019pose, gao2020pose, wang2020high, wang2022pose, miao2021identifying} and human segments~\cite{huang2020human, cai2019multi}. 
%
%
%
%
Another approach is based on feature reconstruction. Hou \etal~\cite{hou2021feature} locate occluded human parts by key-points and propose Region Feature Completion (RFC) to recover the semantics of occluded regions in feature space.
Xu \etal~\cite{xu2022learning} extract occluded features with pose information and propose Feature Recovery Transformer (FRT) to recover the occluded features using the visible semantic features in the k-nearest gallery neighbors. 
%
%
%
In contrast, our approach adaptively removes interference from occluders and background according to correlation within
the class token attention, which generalizes better to various occlusion situations.

\begin{figure*}[t]
\begin{center}
\includegraphics[width=1.0\textwidth]{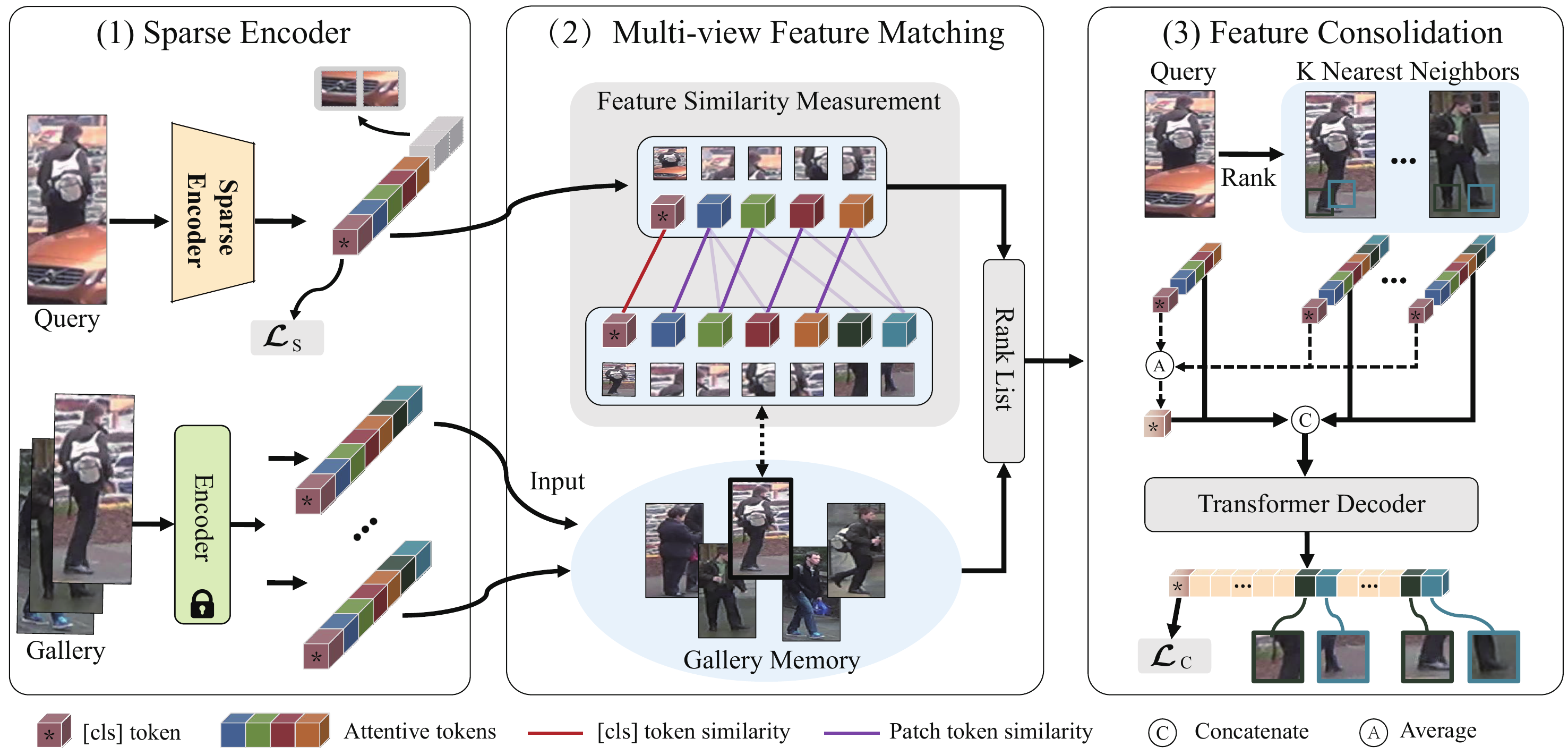}
\end{center}
\caption{Overview of the proposed framework, which consists of a sparse encoder $\mathcal{S}$, a multi-view feature matching module $\mathcal{M}$, and a feature consolidation module $\mathcal{C}$.
The sparse encoder $\mathcal{S}$ removes interfering tokens while preserving attentive tokens. 
In matching module $\mathcal{M}$, we generate a rank list between the sparse query feature and holistic features in a gallery memory pretrained with a vision transformer. We use the summation of image-level cosine distance and patch-level earth mover’s distance as the metric for ranking.
In the consolidation module $\mathcal{C}$, we select the k-nearest neighbors for the query and construct multi-view features by concatenating averaged $[\mathrm{cls}]$ tokens and respective patch tokens of both the query and its selected neighbors. The multi-view features are sent to the transformer decoder for feature consolidation.}
\label{figpipeline}
\end{figure*}

\textbf{Transformer Sparsification}. 
The techniques of accelerating vision transformer models are necessary with limited computing resources. Most methods mainly aim to simplify the structure of the model with efficient attention mechanisms~\cite{Wang2020Linfomer,Kitaev2020Refomer,zhang2022mninivit,zhu2021vtp} and compact structures~\cite{liu2021swin,touvron2021deit,wu2021cvit,graham2021levit}. There are also many researches~\cite{kong2022spvit, rao2021dvit, liang2022evit} focus on effective token learning to reduce the redundancy.
%
%
However, the above approaches have not explored the possibility of applying the characteristics of model acceleration to the occluded person ReID problem. 
Actually, since there are various occlusion and background noise in occluded person Re-ID tasks, we observe that it is sub-optimal to apply the transformer model directly. Following~\cite{liang2022evit}, the purpose of our approach is to prune out ineffective tokens by exploring the sparsity of informative image patches.


\begin{figure}[htbp]
	\centering
    \includegraphics[width=0.42\textwidth]{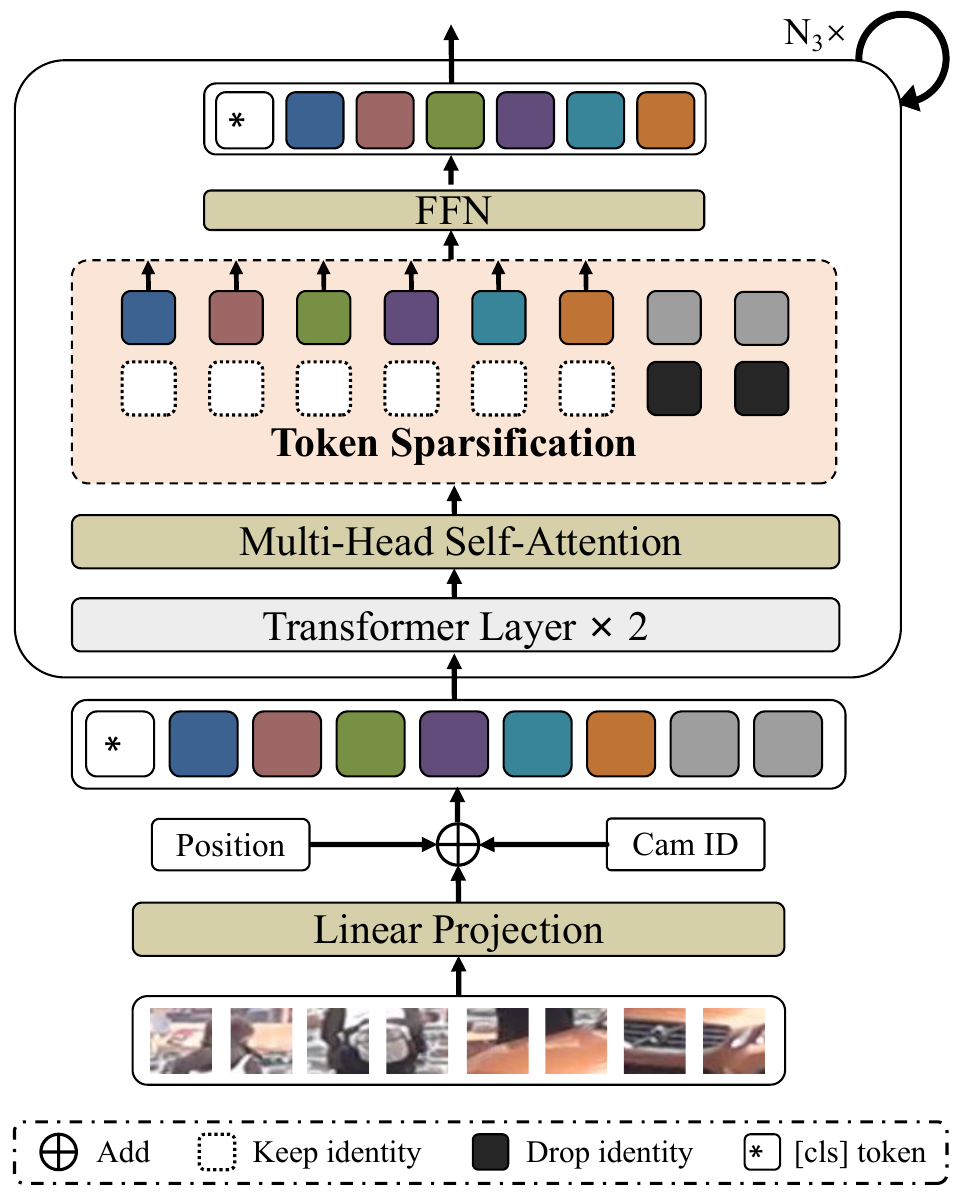} 
	\caption{The structure of our sparse encoder. We divide and embed the image tokens using linear projection, positional encoding, and camera ID encoding. We perform token sparsification in layer 3, 6, and 9.}
    \label{fig:sparse-encoder}
\end{figure}

\section{Method}
We illustrate our feature pruning and consolidation framework as illustrated in~\cref{figpipeline}.
The framework consists of (1) sparse encoder $\mathcal{S}$ conducts token sparsification to prune interference tokens and preserve attentive tokens; (2) multi-view feature matching module $\mathcal{M}$ generates a rank list between the sparse query feature and pre-trained gallery memory by the image and patch-level combined similarity; (3) feature consolidation framework $\mathcal{C}$ utilizes complete information of identified neighbors to compensate pruned query features.

\subsection{Sparse Encoder}
\label{sec:sparse-encoder}
Inspired by the advances in feature pruning~\cite{liang2022evit} and person ReID~\cite{he2021transreid}, we use a transformer with token sparsification to prune interference from occlusion and background.  
As shown in \cref{fig:sparse-encoder}, given an input image $ x\in \mathbb{R} ^{H\times W \times C}$, where $W,H,C$ denote the width, height, and channel of the image respectively, we split $x$ into N overlapping patches $\{p_1, p_2, \cdots , p_N\} $ and embed each patch with linear projection denoted as $f(\cdot)$.
Then we combined a learnable $[\mathrm{cls}]$ token with patch embedding and apply positional encoding and camera index encoding following~\cite{he2021transreid}. The final input can be described as:
\begin{equation}
\mathcal Z_{} = \left \{ x _{cls}, f\left ( p_1 \right ), \cdots ,f\left ( p_N \right )  \right\}  + \mathcal P + \mathcal C_{id}    \label{imgtoken}
\end{equation}
where $x_{cls} \in \mathbb{R}^{1 \times D}$ is learnable $[\mathrm{cls}]$ token. $f(p_{i}) \in \mathbb{R}^{1 \times D}$ is i-th patch embeddings. $\mathcal P\in \mathbb{R} ^{(N+1)\times D}$ is position embeddings and $\mathcal C_{id}\in \mathbb{R} ^{(N+1)\times D}$ is camera index embeddings.

\textbf{Token Sparsification.} 
We adopt the token sparsification strategy proposed in~\cite{liang2022evit}. Specifically, through the attention correlation~\cite{vaswani2017attention} between $[\mathrm{cls}]$ token and other tokens in the vision transformer, we can express the value of $[\mathrm{cls}]$ token as:
\begin{equation}
    x_{cls} = A_{cls} V=\mathrm{softmax}(\frac{Q_{cls} K}{\sqrt{d}}) V \label{qkv}
\end{equation}
where $A_{cls}$ denotes the attention matrix of the $[\mathrm{cls}]$ token, \ie first row of the attention matrix, $\sqrt{d}$ is scale factor. $Q_{cls}, K, V$ represent the query matrix of $[\mathrm{cls}]$ token, the key matrix, and the value matrix respectively.
For multiple heads in the self-attention layer, we average the attention matrix as $\bar{A}_{cls} =  {\textstyle \sum_{i}^{n} \frac{1}{n} \cdot  A_{cls}^{(i)}}$, n is the total heads number.
Since the $[\mathrm{cls}]$ token corresponds to a larger attention value in significant patch regions~\cite{caron2021emerging}, we can evaluate the importance of a token according to its relevance to the $[\mathrm{cls}]$ token. As $\bar{A_{cls}}$ represents the correlation between $[\mathrm{cls}]$ token and all other tokens, we hence preserve the tokens with $\mathcal{K}$ largest values in $\bar{A_{cls}}$ and drop others, which is shown in \cref{fig:sparse-encoder}. We define $ \mathcal{K} = \left \lceil \gamma \cdot N_c  \right \rceil$, where $\gamma$ is the keep rate and $N_c$ is the total number of tokens in the current layer, $\left \lceil  \cdot \right \rceil$ is the ceiling operation.
With token sparsification, the preserved tokens are mostly related to the region of the target pedestrian, and dropped tokens are related to occluders or backgrounds.

\textbf{Sparse Encoder Supervision Loss.} 
We use the cross-entropy ID loss and triplet loss to supervise $[\mathrm{cls}]$ token $x_{cls}^{\mathcal S}$ obtained by the sparse encoder as:
\begin{equation}
    \mathcal L_{S} = \mathcal L_{ID}(x_{cls}^{\mathcal S}) + \mathcal L_{T}(x_{cls}^\mathcal S)
\end{equation}

where $\mathcal L_{ID}$ denotes cross-entropy ID loss and $\mathcal L_{T}$ denotes triplet loss.
Compared with existing approaches, our feature pruning with token sparsification is adaptive and doesn't rely on prior knowledge of human semantics, and can better handle challenging scenarios, such as heavy occlusion and other persons as occluders, as shown in \cref{fig:patch-drop-layer}. 

\subsection{Multi-view Feature Matching Module}
\label{sec:Ranking Strategy}
After the feature pruning, we would like to find the most related patches from other views that not been occluded for consolidation, such strategy has been proved to be effective~\cite{xu2022learning, yu2021neighbourhood}. Specifically, we first learn a gallery memory with the pre-trained encoder. With the pruned query image features, we rank patches in the gallery memory according to their similarity with the query. Considering that there exists appearance gap between the pruned query feature and holistic feature in gallery memory, we measure the similarity from both the image-level and patch-level.
As shown in \cref{figpipeline}, we linearly combine image-level and patch-level distance to match the query image with gallery memory images.

\textbf{Image-level distance.}
We define the image-level distance as the cosine similarity between the $[\mathrm{cls}]$ tokens of query and gallery memory as follows:
\begin{equation}
\mathcal D_{COS} =  1 - \frac{  \langle x_{cls}^{(i)}, x_{cls}^{(i)}  \rangle }{ | x_{cls}^{(i)}  |\cdot  |  x_{cls}^{(j)}  | } \label{cos-dist}
\end{equation}
where $x_{cls}^{(i)}$ and $x_{cls}^{(j)}$ is $[\mathrm{cls}]$ token of i-th query image and j-th image in gallery memory. $\left \langle \  \cdot \  \right \rangle $ is the dot product.

\textbf{Patch-level distance.}
We leverage Earth Mover’s Distance (EMD)~\cite{rubner2000earth} to measure patch-level similarity. EMD is usually employed for measuring the similarity between two multidimensional distributions and is formulated as following linear programming problem: Let $\mathcal Q = \left \{ (q_1, w_{q_1}), \dots ,(q_m, w_{q_m}) \right \} $ be the set of patch tokens from the query image, where $q_i$ is i-th patch token and $w_{q_i}$ is the weight of $q_i$. Similarly, $\mathcal G = \left \{ (g_1, w_{g_1}), \dots ,(g_n, w_{g_n}) \right \} $ represents set of patch tokens of a gallery image.
Then $w_{q_i}$ is further defined as proportional correlation weight~\cite{phan2022deepface} of $q_i$ to set $\{g_1, g_2, \dots , g_n\}$, denoted as $ \max (0, \langle q_i,\frac{1}{n} \sum_{i}^{n}g_i \rangle )$, and $w_{g_j}$ equals $\max (0, \langle g_j,\frac{1}{m} \sum_{i}^{m}q_i \rangle )$. The ground distance $d_{ij}$ is the cosine distance between $p_i$ and $q_j$.
The objective is to find the flow $\mathcal{F}$, of which $f_{ij}$ indicates the flow between $p_i$ and $q_j$, to minimize following cost:
\begin{equation}
\begin{split}
    \mathcal F^{*} & = \operatorname*{arg\,min}_{\mathcal F} \mathrm{COST} (\mathcal Q, \mathcal G, \mathcal F)\\
    & = \operatorname*{arg\,min}_{\mathcal F} \left ( \sum_{i=1}^{m} \sum_{j=1}^{n}f_{ij}d_{ij} \right ) \label{emdcost}
\end{split}
\end{equation}    

subject to the following constraints: 
\begin{equation}
    f_{ij} \geq 0, \ 1 \le i\le m, \ 1\le j\le n
\end{equation}
\begin{equation}
    \sum_{i=1}^{m} f_{ij} \le w_{g_j}, \ 1 \le j \le n
\end{equation}
\begin{equation}
    \sum_{j=1}^{n} f_{ij} \le w_{q_i}, \ 1 \le i \le m 
\end{equation}
\begin{equation}
    \sum_{i=1}^{m} \sum_{j=1}^{n}f_{ij} = \min (\sum_{i=1}^{m} w_{q_i}, \sum_{j=1}^{n} w_{g_j})
\end{equation}

\cref{emdcost} can by solved by iterative Sinkhorn algorithm~\cite{cuturi2013sinkhorn} and produces the earth mover's distance $\mathcal{D}_{EMD}$. Intuitively, $\mathcal{D}_{COS}$ represents global distance, while $\mathcal{D}_{EMD}$ measures in the perspective of the set of local features. Naturally, we take the linear combination of image-level cosine distance and patch-level EMD as our final distance for feature matching, expressed as follows:

\begin{equation}
    \mathcal{D} =(1 -\alpha) \mathcal D_{COS} + \alpha  \mathcal D_{EMD}
    \label{eq:distance weight}
\end{equation}

With $\mathcal{D}$, we rank the sparse query feature with holistic features in gallery memory and generate a ranking list.
In order to expedite the process of feature matching, we use an efficient two-stage selection strategy to save time costs by 138 times. 

%

\subsection{Feature Consolidation Decoder}
\label{sec:Feature Recover Decoder}
After we retrieve the $K$ nearest gallery neighbors, we then consolidate the pruned query feature from multi-view observations in the gallery memory.
Specifically, we average the $[\mathrm{cls}]$ tokens of the query and gallery neighbors as our initial global feature. Then we combine the averaged $[\mathrm{cls}]$ token with patch tokens in both query and gallery neighbors to aggregate information of multi-view pedestrians. The consolidation of the multi-view features can be formulated as follows:
\begin{equation}
f_{m} = \left \{  \bar {f_c}, f_{p_q}, f_{p_g}^{{(1)}}, f_{p_g}^{(2)}, \dots, f_{p_g}^{(K)}  \right \} 
\end{equation}
where $\bar {f_c} \in \mathbb{R}^{1 \times C}$ is the average of $\mathrm{[cls]}$ tokens of both query and its neighbors. $f_{p_q} \in \mathbb{R}^{M \times C}$ is the patch tokens of query and $M$ is the number of patch tokens. $f^{(i)}_{p_g} \in \mathbb{R}^{N \times C}$ is the patch tokens of i-th gallery neighbor and $N$ is the number of patch tokens for each gallery.

\textbf{Transformer Decoder.}
Notice that $[\mathrm{cls}]$ token is class prediction~\cite{dosovitskiy2020image} in transformer and conventionally treated as global feature for image representation, the query $\mathrm{[cls]}$ token is usually contaminated in the occluded person ReID problem . 
We send the consolidated multi-view feature to a transformer decoder to compensate the incomplete $\mathrm{[cls]}$ token with gallery neighbors. In the decoder, we first transform the multi-view feature $f_{m}$ into $Q, K, V$ vectors using linear projections, which is:
\begin{equation}
Q=W_q \cdot f_m, K=W_k \cdot f_m, V=W_v \cdot f_m
\end{equation}
where $W_q, W_k, W_v$ are weights of linear projections. As the multi-view feature contains three parts: $\mathrm{[cls]}$ token, patch tokens of query, and patch tokens of gallery neighbors, the computation of attention with respect to $\mathrm{[cls]}$ token can be decomposed into the above three parts, which is expressed as:
\begin{equation}
\begin{split}
x_{cls} &= \mathrm{Cat}(A^{'}_c, A^{'}_q, A^{'}_g) \cdot \mathrm{Cat}(V_{c}, V_{q}, V_g) \\
            &= A^{'}_c \cdot V_c + A^{'}_q \cdot V_q + A_g^{'} \cdot V_g
\end{split}
\label{consolidation-cls-otken}
\end{equation}
where $\mathrm{Cat} \left ( \  \cdot \  \right ) $ is operation of vector combination.
$A^{'}$ denotes the attention matrix of $\mathrm{[cls]}$ token and $V$ is the value vector. Subscripts $c, q, g $ correspond to $\mathrm{[cls]}$ token, query and gallery neighbors respectively. $A_c^{'} \cdot V_c + A_q^{'} \cdot V_q$ acts as the feature learning process with sparse query and $A_g^{'} \cdot V_g$ integrates completion information from gallery neighbors to the $\mathrm{[cls]}$ token. 
%
%
The final consolidated $[\mathrm{cls}]$ token generated by the transformer decoder is denoted as $x_{cls}^{\mathcal C} = \tau (f_m)$. We find one layer of transformer is enough and also avoids the high memory consumption from neighborhoods~\cite{yu2021neighbourhood}.

\textbf{Consolidation Loss.} 
The $[\mathrm{cls}]$ token $x_{cls}^{\mathcal C}$ obtained by feature consolidation decoder is supervised with cross entropy loss as ID loss and triplet loss, as expressed below:
\begin{equation}
    \mathcal L_{C} = \mathcal L_{ID}(x_{cls}^{\mathcal C}) + \mathcal L_{T}(x_{cls}^\mathcal C)
\end{equation}
Therefore, the final loss function can be expressed:
\begin{equation}
    \mathcal L = \mathcal L_{S} + \mathcal L_{C}
\end{equation}

\subsection{Implementation Details}
\label{implementation}
{\color{black} We choose the ViT-B/16 as our backbone for both the sparse encoder and gallery encoder. 
Our sparse encoder incorporates several modifications into the backbone, including camera encoding, patch overlapping with a stride of 11, batch normalization, and token sparsification at layer 3, 6, 9. 
Our gallery encoder has the same structure as the sparse encoder but does not conduct token sparsification. We construct the gallery memory with gallery image tokens as in \cref{imgtoken}.}
In the multi-view feature matching module, we first identify 100 globally-similar gallery neighbors using cos distance in \cref{cos-dist}, and then search the final $K$ nearest gallery neighbors using the proposed distance in \cref{eq:distance weight}.
We set the number of $K$ to 10 on Occluded-Duke dataset and 5 for others.
In the training process, we resize all input images to 256 $\times$ 128. The training images are augmented with random horizontal flipping, padding, random cropping and random erasing~\cite{zhong2020random}. The batch size is 64 with 4 images per ID and the learning rate is initialized as 0.008 with cosine learning rate decay. The distance weight $\alpha$ in \cref{eq:distance weight} is 0.4 and the parameter of keep rate $\gamma$ is 0.8.
We take consolidated $\mathrm{[cls]}$ token in \cref{consolidation-cls-otken} for model inference.
%
For large-scale problems, we can further replace the full image tokens with [cls] tokens to save computation and memory.

\section{Experiment}

We conduct extensive experiments to validate our framework. We first introduce the dataset we use:
\subsection{Datasets and Evaluation Metric}
\textbf{Occluded-Duke}~\cite{miao2019pose} includes 15,618 training
images of 702 persons, 2,210 occluded query images of 519 persons,
and 17,661 gallery images of 1,110 persons.
\textbf{Occluded-ReID}~\cite{zhuo2018occluded} consists of 1,000 occluded query images and 1,000 full-body gallery images both belonging to 200 identities.
\textbf{Partial-ReID}~\cite{zheng2015partial} involves 600 images from 60 persons, and each person consists 5 partial and 5 full-body images.
\textbf{Market-1501}~\cite{zheng2015scalable} contains 12,936 training images of 751 persons, 19,732 query images and 3,368 gallery images of 750 persons captured by six cameras.


\textbf{Evaluation Metirc}. All methods are evaluated under the Cumulative Matching Characteristic (CMC) and mean Average Precision (mAP). Floating Point Operations (FLOPs) represents the amount of model computation.

\begin{table}[t]
\centering
\begin{tabular}{lcccc}
\hline
                         & \multicolumn{2}{c}{Occluded-Duke}     & \multicolumn{2}{c}{Occluded-ReID} \\
\multirow{-2}{*}{Method} & Rank-1               & mAP          & Rank-1       & mAP          \\ \hline\hline
DSR                       & 40.8                & 30.4         & 72.8         & 62.8         \\
PGFA                      & 51.4                & 37.3         & -            & -            \\
HOReID                    & 55.1                & 43.8         & 55.1         & 43.8         \\
OAMN                      & 62.6                & 46.1         & -            & -            \\
PAT                       & 64.5                & 53.6         & 81.6         & 72.1         \\
TransReID                 & 67.4                & 59.5         & -            & -            \\
PFD                       & 69.5                & 61.8         & 81.5         & 83.0         \\
FED                       & 67.9                & 56.3         & \textbf{87.0}& 79.4         \\ 
RFCnet                    & 63.9                & 54.5         & -            & -            \\
Yu et al.                 & 67.6                & 64.2         & 68.8         & 67.3         \\
FRT                       & 70.7                & 61.3         & 80.4         & 71.0         \\ \hline
FPC (ours)                & \textbf{76.7}       & \textbf{72.8}& 86.3         & \textbf{84.6}\\ \hline
\end{tabular}
\caption{Comparison with state-of-the-art methods on Occluded-Duke and Occluded-ReID datasets.}
\label{table:occluded-ReID-dataset}
\end{table}

\begin{table}[t]
\centering
\begin{tabular}{lcccc}
\hline
                         & \multicolumn{2}{c}{Market-1501}       & \multicolumn{2}{c}{Partial-ReID} \\
\multirow{-2}{*}{Method} & Rank-1               & mAP          & Rank-1       & mAP          \\ \hline\hline
PCB                      & 92.3                & 77.4         & -            & -            \\
PGFA                     & 91.2                & 76.8         & 69.0         & 61.5         \\
HOReID                   & 94.2                & 84.9         & 85.3         & -            \\
OAMN                     & 92.3                & 79.8         & 86.0         & -            \\
PAT                      & 95.4                & 88.0         & 88.0         & -            \\
TransReID                & 95.0                & 88.2         & 83.0         & 77.5         \\
PFD                      & \textbf{95.5}       & 89.7         & -            & -            \\
FED                      & 95.0                & 86.3         & 84.6         & 82.3         \\ 
RFCnet                   & 95.2                & 89.2         & -            & -            \\
Yu et al.                & 94.5                & 86.5         & -            & -            \\
FRT                      & \textbf{95.5}       & 88.1         & \textbf{88.2}& -            \\ \hline
FPC (ours)               & 95.1                & \textbf{91.4}& 86.3         & \textbf{86.5}\\ \hline
\end{tabular}
\caption{Comparison with state-of-the-art methods on Market-1501 and Partial-ReID datasets.}
\label{table:holistic-dataset}
\end{table}

\subsection{Comparison with State-of-the-art Methods}
\textbf{Experimental results on Occluded ReID Datasets.}
In \cref{table:occluded-ReID-dataset}, we compare FPC with state-of-the-art methods on two occluded ReID datasets (\ie Occluded-Duke and Occluded-ReID). 
{\color{black} Methods in different categories are compared, including the partial ReID methods~\cite{he2018deep}, key-points based methods~\cite{miao2019pose, wang2020high, wang2022pose, hou2021feature}, data augmentation methods~\cite{chen2021occlude, wang2020high, wang2022feature}, transformer-based methods~\cite{li2021diverse, he2021transreid}, gallery-based reconstruction methods~\cite{ yu2021neighbourhood, xu2022learning}. In addition, TransReID~\cite{he2021transreid} and PFD~\cite{wang2022pose} use camera information.}
Since no training set is provided for Occluded-ReID, we adopt Market-1501 as the training set the same as other methods to ensure the fairness of comparison.
We can see that our FPC outperforms existing approaches and demonstrates the effectiveness for the occluded ReID tasks.
Specifically, our FPC achieves the best performance on the challenging Occluded-Duke dataset, outperforming other methods by at least 6.0\% Rank-1 accuracy and 8.6\% mAP.
On the Occluded-ReID dataset, FPC produces the highest mAP, outperforming the other methods by at least 1.6\%. FPC achieves comparable results in Rank-1 accuracy with FED~\cite{wang2022feature}, and much better than others.


\textbf{Experimental Results on Holistic and Partial ReID Datasets.}
Many existing occluded ReID methods can not be effectively applicable to holistic and partial ReID datasets. On the contrary, FPC achieves great performance improvement on holistic and partial datasets, \ie Market-1501 and Partial-ReID. The results are shown in~\cref{table:holistic-dataset}.
We compare FPC with three categories of methods, including the holistic ReID methods~\cite{sun2018beyond}, the current leading methods in occluded ReID~\cite{miao2019pose, wang2020high, chen2021occlude, li2021diverse, he2021transreid, wang2022pose, wang2022feature}, the feature reconstruction based methods in occluded ReID~\cite{hou2021feature, yu2021neighbourhood, xu2022learning}.
On the Partial-ReID dataset, FPC achieves the best mAP result, lower than the PAT and FRT in Rank-1 accuracy. We consider that we adopt the ViT models and are more prone to overfit on small datasets which leads poor cross-domain generalization.
We also observe that FPC achieves competitive Rank-1 accuracy and the highest mAP on the Market-1501 dataset, which is at least 1.7\% mAP ahead of other methods.
The excellent performance on holistic and partial ReID datasets illustrates the robustness of our FPC.


\begin{table}[t]
\centering
\begin{tabular}{cccccc}
\hline
Index & $\mathcal S$ & $\mathcal M$ & $\mathcal{C}$ & Rank-1 & mAP  \\ \hline
1     & \ding{56}  & \ding{56}   & \ding{56}   & 65.2 & 56.6 \\
2     & \ding{52}  & \ding{56}   & \ding{56}   & 68.6 & 60.1 \\
3     & \ding{52}  & \ding{56}   & \ding{52}   & 75.8 & 71.6 \\
4     & \ding{52}  & \ding{52}   & \ding{52}   & 76.7 & 72.8 \\ \hline
\end{tabular}
\caption{Ablation study on each component.}
\label{table:modules}
\end{table}

\begin{table*}[htbp]
\centering
\begin{tabular}{ccccccl}
\hline
\multirow{2}{*}{Metrics} & \multicolumn{6}{c}{Keep Rate $\gamma$}           \\ \cline{2-7} 
                         & 1.0  & 0.9  & 0.8  & 0.7  & 0.6  & 0.5  \\ \hline
Rank-1 (\%)                   & 76.0 & 76.6 {\color{red} (+0.6)} & 76.7 {\color{red} (+0.7)} & 75.9 {\color{red} (-0.1)} & 74.5 {\color{red} (-1.5)} & 72.1 {\color{red} (-3.9)} \\ 
mAP (\%)                      & 72.3 & 72.7 {\color{red} (+0.4)} & 72.8 {\color{red} (+0.5)} & 72.4 {\color{red} (+0.1)} & 71.3 {\color{red} (-1.0)} & 69.3 {\color{red} (-3.0)} \\ 
FLOPs ($\mathrm{G}$)                   & 20.8 & 18.1 {\color{red} (-13\%)} & 15.7 {\color{red} (-25\%)} & 13.6 {\color{red} (-33\%)} & 11.9 {\color{red} (-43\%)} & 10.4 {\color{red} (-50\%)} \\
\hline
\end{tabular}
\caption{Analysis on effectiveness of the keep rate $\gamma$. We perform FLOPs comparison on the sparse encoder.}
\label{table:flops}
\end{table*}

\subsection{Ablation Study}
\label{Ablation}

\textbf{Analysis of proposed components.}
The results are shown in \cref{table:modules}. 
{\color{black} Index-1 is our baseline architecture. To assess the effectiveness of $\mathcal{S}$, we conduct a comparison between index-1 and index-2. Our findings demonstrate that $\mathcal{S}$ leads to a 3.4\% improvement in Rank-1 accuracy and a 3.5\% increase in mAP over the baseline. This suggests that $\mathcal{S}$ has the potential to mitigate the interference of inattentive features (mainly related to occlusion and background noise). Furthermore, $\mathcal{S}$ can also expedite model inference, as elaborated in \cref{table:flops}.}
%
%
By comparing index-2 and index-3, $\mathcal{C}$ improves performance by 7.2\% Rank-1 accuracy and 11.5\% mAP, which indicates that $\mathcal{C}$ can effectively compensate the occluded query feature and bring huge performance improvements.
By comparing index-3 and index-4, $\mathcal{M}$ can increase performance by 0.9\% Rank-1 accuracy and 1.2\% mAP, indicating that the proposed distance metric contributes to the find the accurate gallery neighbors.

\begin{figure}[t]
    \centering
    \begin{subfigure}{0.495\linewidth}
        \includegraphics[width=\linewidth]{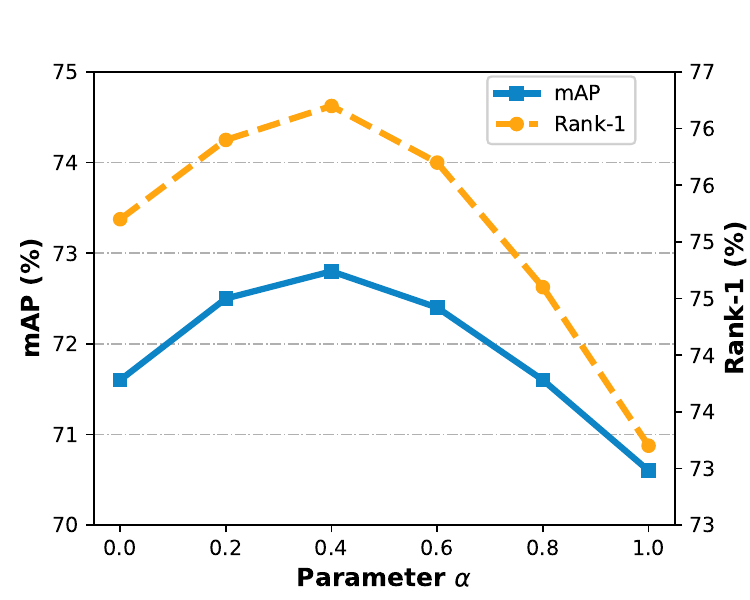}
        \caption{}
        \label{fig:paras-a}
    \end{subfigure}
    \begin{subfigure}{0.495\linewidth}
        \includegraphics[width=\linewidth]{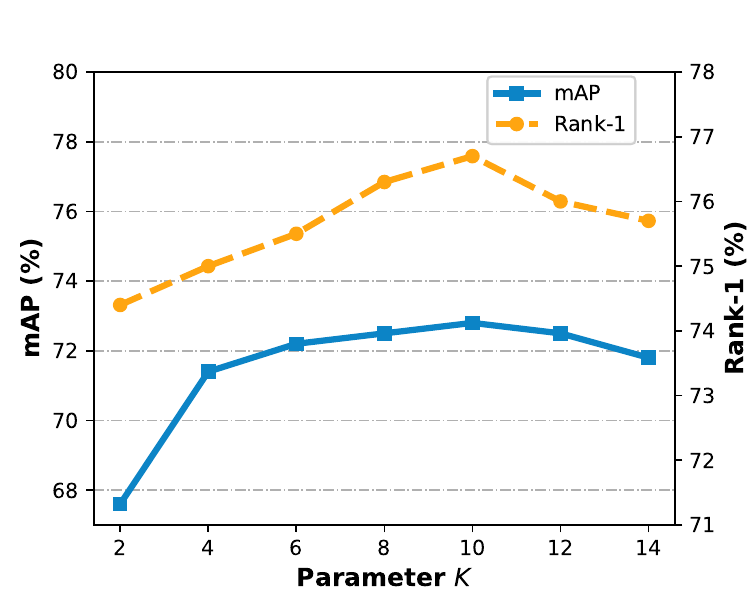}
        \caption{}
        \label{fig:paras-b}
    \end{subfigure}
    \caption{Analysis of distance weight $\alpha$ and the number of nearest gallery neighbors $K$ on Occluded-Duke dataset.}
    \label{fig:paras}
\end{figure}


\textbf{Analysis of Distance Weight $\alpha$}. The distance weight $\alpha$ defined in \cref{eq:distance weight} balances the importance between image-level and patch-level distance.
From \cref{fig:paras-a}, as $\alpha$ goes from 0 to 1, the patch-level distance gradually takes effect. When $\alpha$ is 0.4, the combination of both achieves the best performance with 76.7\% Rank-1 accuracy and 72.8\% mAP. 

\textbf{Analysis of Number $K$ of Nearest Gallery Neighbors}.
We conduct quantitative experiments to find the most appropriate number $K$ of nearest gallery neighbors. 
We conclude from \cref{fig:paras-b} that too small a choice of $K$ lacks sufficient information for feature consolidation and too large a choice increases the risk of incorrectly selecting neighbors.
When the value of $K$ is 10, we achieve the optimal performance.


\begin{figure}[htbp]
	\centering
	\includegraphics[width=0.42\textwidth]{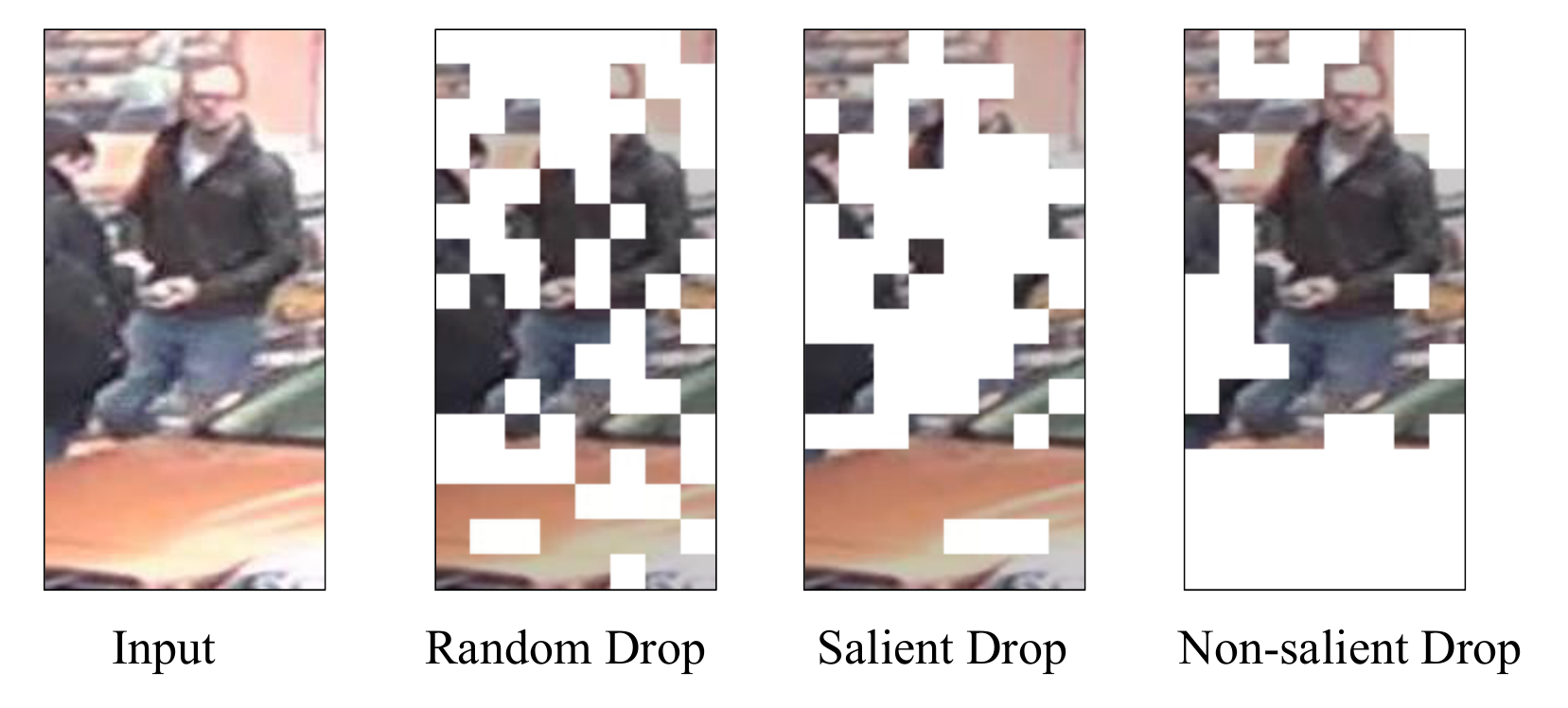} 
	\caption{Illustration of three different patch drop strategies. White patches are the dropped image parts.}
	\label{fig:patch-drop}
\end{figure}

\begin{figure}[htbp]
    \centering
    \begin{subfigure}{0.495\linewidth}
        \includegraphics[width=\linewidth]{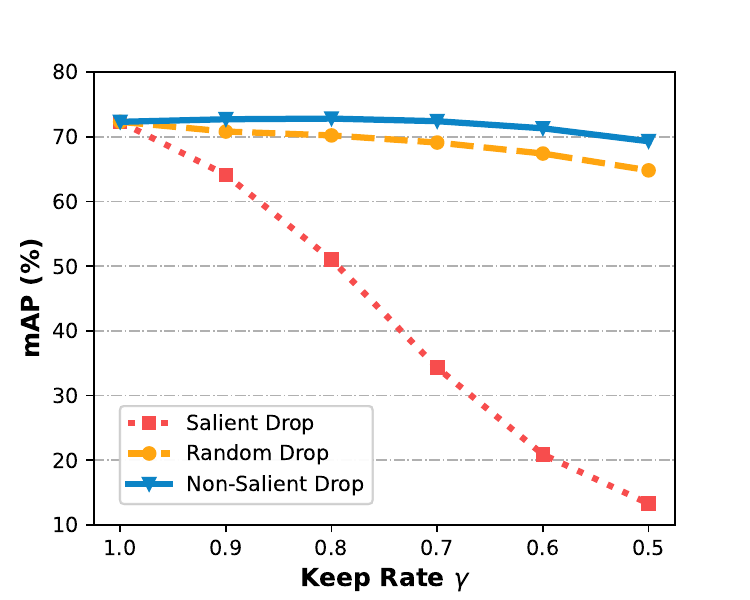}
    \end{subfigure}
    \begin{subfigure}{0.495\linewidth}
        \includegraphics[width=\linewidth]{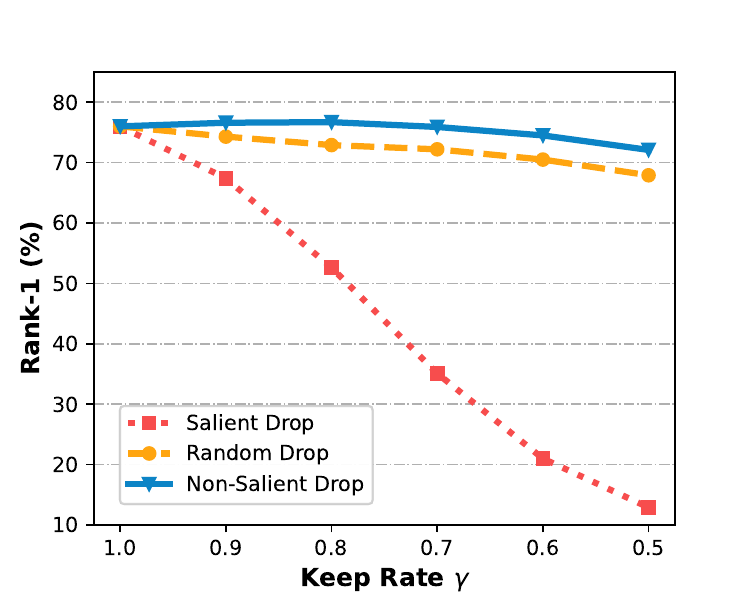}
    \end{subfigure}
    \caption{The analysis of three different patch drop strategies.}
    \label{fig:different patch drop}
\end{figure}

\textbf{Analysis of Keep Rate $\gamma$ in Sparse Encoder.}
Here, keep rate $\gamma$ reflects the number of preserved tokens in each layer of the sparse encoder.
We find instructive observations from \cref{table:flops}: as the $\gamma$ goes from 1.0 to 0.8, the FLOPs drops while the model performance improves.
This suggests that a reasonable choice of $\gamma$ can effectively filter out inattentive features, thus reducing the computational complexity and enhancing the model inference capability.
When $\gamma$ is 0.8, we achieve the best balance between computational complexity and experimental performance, which leads to an improvement of 0.5\% mAP, 0.7\% Rank-1 accuracy and 25\% reduction in FLOPs.

\textbf{Analysis of Patch Drop Strategy in Sparse Encoder.}
We experiment with three variants of patch drop approaches to demonstrate the effectiveness of the proposed method. 
As shown in \cref{fig:patch-drop}, we preserve the same number of patches and choose different preservation methods: 
Random Drop means that $\mathcal{K}$ patches are randomly selected to be retained. Non-salient Drop, as the proposed method, preserves the patches corresponding to the $\mathcal{K}$ largest values in the $\mathrm{[cls]}$ token attention.
Conversely, Salient Drop preserves the $\mathcal{K}$ smallest ones.
The performance of the three patch drop approaches with different keep rate is shown in \cref{fig:different patch drop}.
We observe that the proposed Non-salient Drop method achieves best performance under different keep rate settings, indicating the importance of attentive features for feature matching and consolidation.
{\color{black} In addition, the result of random drop test still achieves relatively high performance, which reflects that our proposed sparse encoder structure is robust against information loss.}

\textbf{Analysis of speed and storage optimization.}
 {\color{black} Our approach relies on gallery information for feature consolidation, to address the practical concerns regarding speed and storage for real-world problems, we conduct analysis on our optimization strategies on the Occluded-Duke dataset. 
 In $\mathcal{S}$, token sparsification reduce 25\% Flops (in \cref{table:flops}) with $\gamma$ equals 0.8 and thus accelerates model inference. 
 %
%
In $\mathcal{M}$, we use a two-step matching procedure, i.e., use cosine similarity for initial filtering and compute EMD in the finer step, achieving a remarkable 138-fold reduction in computational costs, compared to the exhaustive search based on EMD. Our approach takes 0.43ms per image, while the full EMD calculation takes 60.0ms per image. 
Furthermore, if we replace full image tokens with $\mathrm{[cls]}$ token, our approach saves a lot memory (\ie only takes 0.05G for the entire gallery set) and achieves 17x faster in nearest neighbor search time (\ie from 0.43ms to 0.026ms per image), while still preserving an acceptable degree of precision (\ie mAP: 72.8\% to 70.2\%, Rank-1: 76.7\% to 74.3\%, which still outperforms others).}

\textbf{Comparisons with Re-ranking.}
{\color{black}
An alternative approach is the re-ranking technique~\cite{zhong2017re}, which also uses k-nearest gallery neighbors information. We compare $\mathcal{C}$ with re-ranking~\cite{zhong2017re}, 
%
the second group results in \cref{table:re-ranking} indicate that re-ranking fails to attain comparable performance of $\mathcal{C}$, lead to a reduction of 3.2\% Rank-1 and 0.3\% mAP.
%
%
Further, our $\mathcal{M} + \mathcal{C}$ is approximately 17 times faster than re-ranking, with respective times of 0.47ms and 7.9ms per image.}
Moreover, the last group shows $\mathcal{C}$ and re-ranking can be jointly employed and outperforms others. 


\begin{table}[t]
\centering
\begin{tabular}{ccc}
\hline
Methods & Rank-1 & mAP  \\ \hline
FRT\cite{xu2022learning}      & 70.7 & 61.3 \\
FRT\cite{xu2022learning} + re-ranking      & 70.8 & 65.0 \\ 
Yu \etal~\cite{yu2021neighbourhood}      & 67.6 & 64.2 \\
Yu \etal~\cite{yu2021neighbourhood} + re-ranking      & 68.9 & 67.3 \\ \hline
$\mathcal S$       & 68.6 & 60.1  \\
$\mathcal S$ + re-ranking      & 72.6 & 71.3  \\ 
$\mathcal S$ + $\mathcal C$      & 75.8 & 71.6  \\ \hline
FPC (ours)      & 76.7 & 72.8 \\ 
FPC (ours) + re-ranking      & \textbf{78.6} & \textbf{78.3} \\ \hline
\end{tabular}
\caption{Performance comparsions with re-ranking on Occluded-Duke dataset. $\mathcal S$ is our sparse encoder and $\mathcal C$ is feature consolidation.}
\label{table:re-ranking}
\end{table}

%

\begin{figure}[t]
	\centering
	\includegraphics[width=0.45\textwidth]{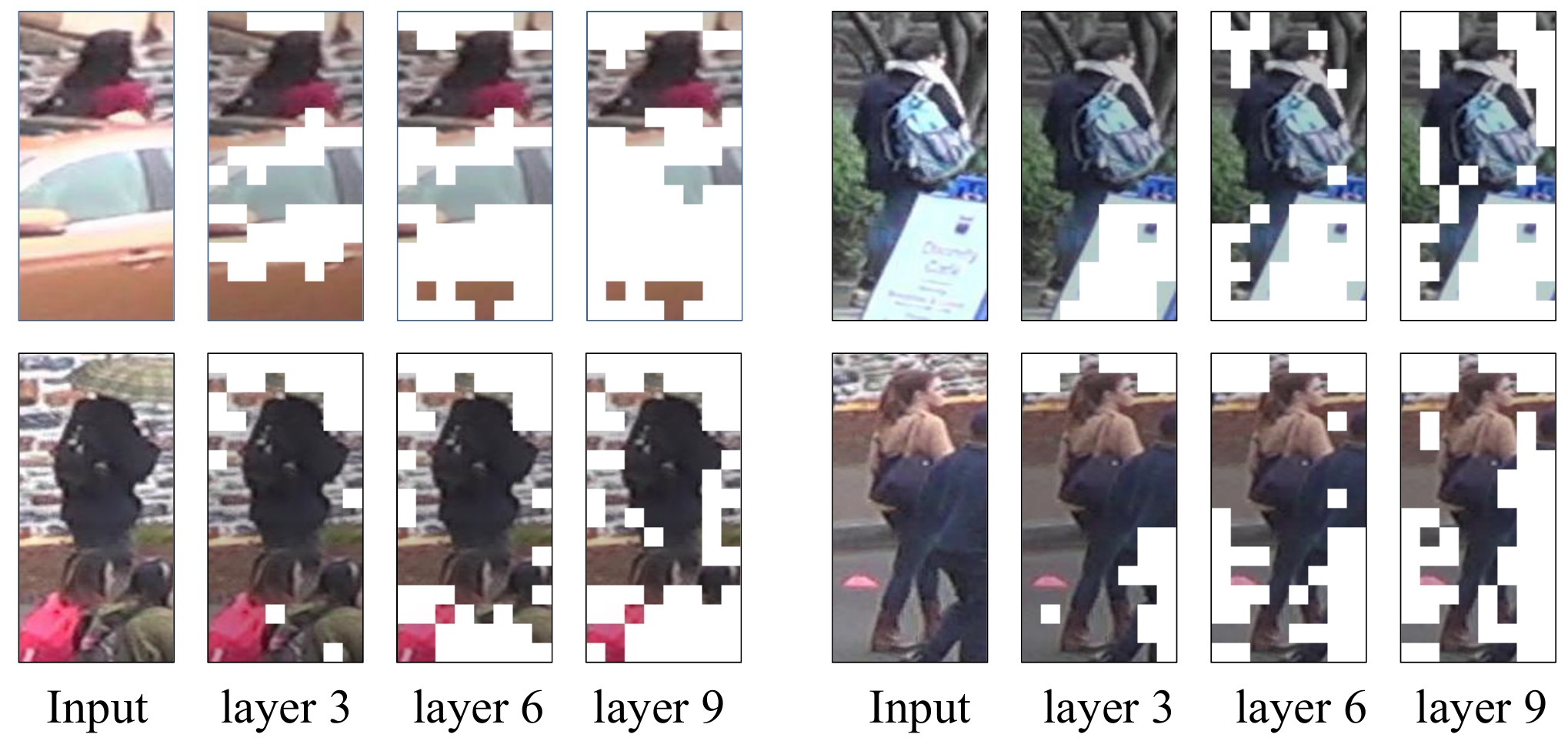} 
	\caption{Visualization of patch drop process in different layers of the sparse encoder. We show various occlusion scenarios such as object occlusion, pedestrian occlusion and heavy occlusions.}
	\label{fig:patch-drop-layer}
\end{figure}


\subsection{Visualization}

\textbf{Patch pruning.}
The patch-drop process in different layers of sparse encode is shown in \cref{fig:patch-drop-layer}. We observe that with the increasing number of sparse encoder layers, more object occlusion, pedestrian occlusion and background noise are filtered out, while
the essential classification information of target pedestrians
is preserved.




\section{Conclusion}
In this paper, we propose the feature pruning and consolidation (FPC) framework for the occluded person ReID task.
Specifically, our framework prunes interfering image tokens (mostly related to background noise and occluders) without relying on prior human shape information. We propose an effective way to consolidate the pruned feature and achieve the SOTA performance on many occluded, partial, and holistic ReID datasets. We introduce the token sparsification technique and demonstrate its effectiveness. 

\newpage

\section{Acknowledgment}
This work is supported by the National Natural Science Foundation of China (NSFC No. 62272184). The computation is completed in the HPC Platform of Huazhong University of Science and Technology.


\bibliography{aaai24}

\newpage
\section{Supplementary Material}
We present additional experimental and qualitative results in this supplementary material. 

\section{More Experimental Details and Results} 
\subsection{Visualization Results with Different Keep Rate}
\begin{figure}[h]
	\centering
	\includegraphics[width=0.45\textwidth]{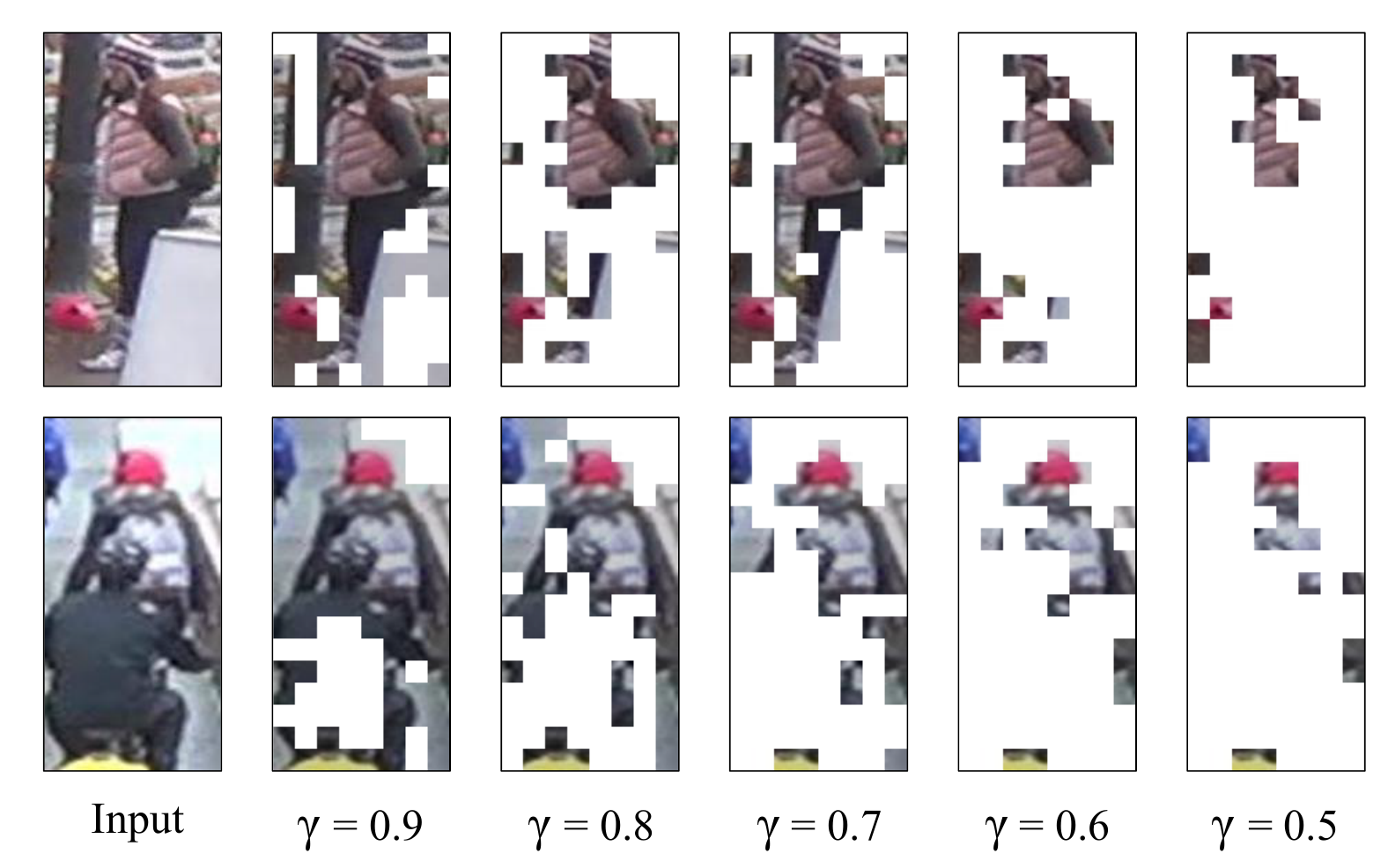} 
	\caption{Visualization of patch drop with different keep rate $\gamma$ in sparse encoder.}
	\label{fig:patch-drop-keeprate}
\end{figure}

We show the visualization of patch drop with different keep rate $\gamma$ in \cref{fig:patch-drop-keeprate}. $\gamma$ reflects the number of preserved patches in the sparse encoder. As $\gamma$ decreases from 0.9 to 0.5, the patches corresponding to backgrounds and occlusions are gradually pruned. When the $\gamma$ is 0.5, only important target human areas and several outliers are preserved.

\subsection{Extended Visualization Results of Sparse Encoder}
\begin{figure*}[t]
    \includegraphics[width=1.0\linewidth]{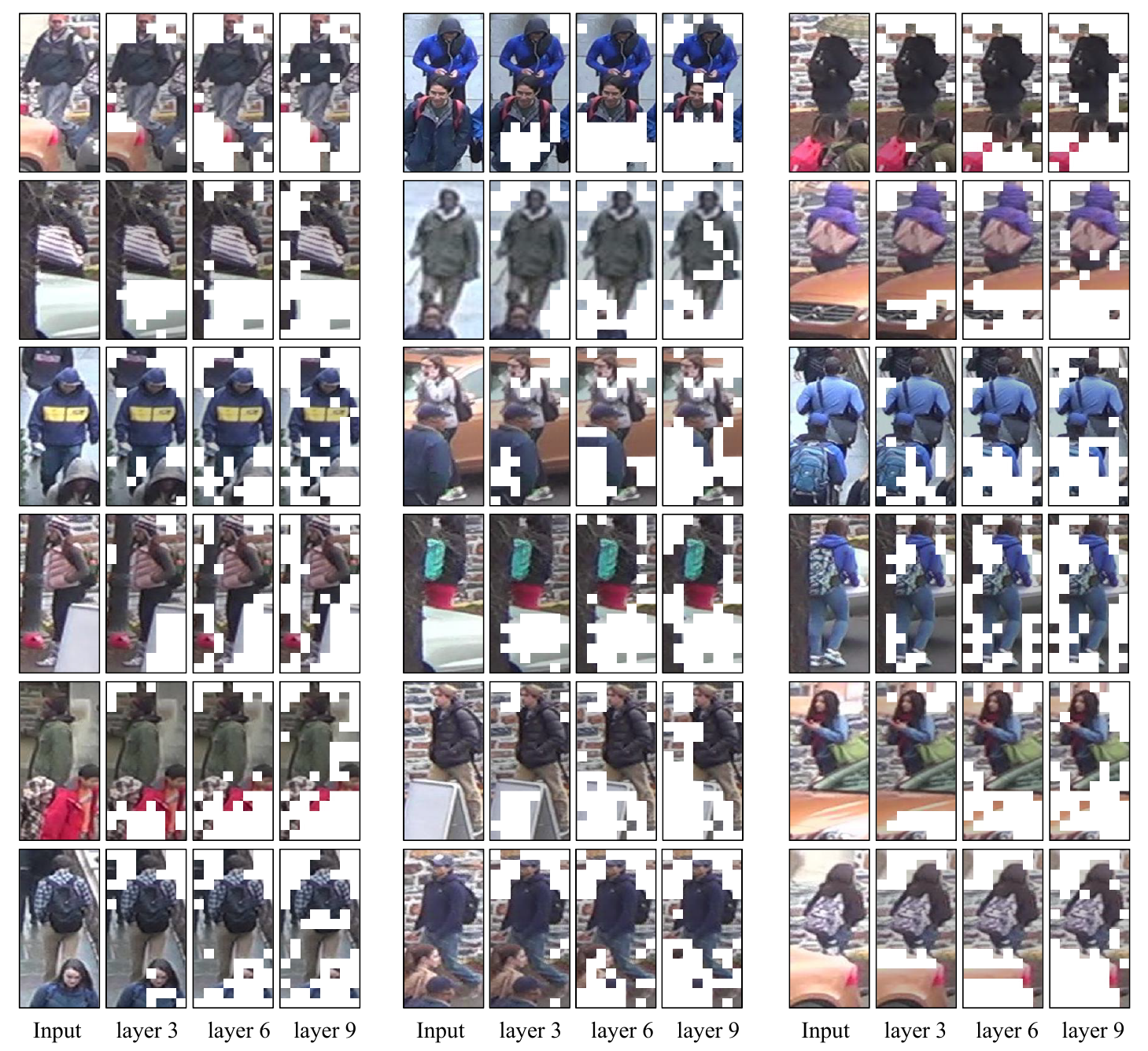} 
	\caption{Extended visualization results of inattentive tokens in different layer of sparse encoder. The regions without masks represent the attentive tokens (mainly relate to target person). The masked regions denote the inattentive tokens (mostly related to background noise and occlusions). Our sparse encoder is effective in dealing with occlusion and background.}
    \label{fig:supp_vis_patch_drop_layer}
\end{figure*}

We show more visualization results in \cref{fig:supp_vis_patch_drop_layer} in order to illustrate the attentive token identification. Input images are chosen from Occluded-Duke\cite{miao2019pose}, including object occlusion and pedestrian occlusion. The results validate the ability of our sparse encoder to be applicable to a variety of occlusion cases.

\subsection{Analysis of Token Sparsification}
In our paper, we discuss the effect of token sparsification, \ie, preserving attentive tokens and dropping inattentive tokens, which can both reduce model computation and enhance model inference.
Similarly, this sparsification mechanism is also applicable to other occluded ReID methods. Here we demonstrate the effectiveness of token sparsification with Transreid \cite{he2021transreid} on Occluded-Duke\cite{miao2019pose} and Occluded-ReID\cite{zhuo2018occluded} datasets, as shown in \cref{table:token sparsification}. We discover that the token sparsification (abbreviated as T.S.) improves the performance of Rank-1 accuracy and mAP on both datasets, which demonstrates that token sparsification may have broad scope of application in occluded ReID tasks.

\begin{table}[h]
\centering
\begin{tabular}{ccccc}
\hline
& \multicolumn{2}{c}{Occluded-Duke} & \multicolumn{2}{c}{Occluded-ReID} \\
\multirow{-2}{*}{Method} & Rank-1 & mAP & Rank-1 & mAP \\ \hline
Transreid  & 67.4 & 59.5 & 80.5 & 74.4  \\ 
Transreid + T.S. & \textbf{68.2} & \textbf{60.0} & \textbf{80.9} & \textbf{75.0} \\ \hline 
\end{tabular}
\caption{Performance comparison with token sparsification on Occluded-Duke and Occluded-ReID datasets. Here we set keep rate to 0.9.}
\label{table:token sparsification}
\end{table}

\subsection{Ablation Study on Layers of Token Sparsification}
We have performed an ablation study on the token sparsification layers, as detailed in ~\cref{layers}. Our results indicated that the foremost layer has a significant impact on performance, and an excessive forward shift of the layer may lead to performance degradation.

\begin{table}[h]
\centering
    \resizebox{1.0\columnwidth}{!}{
    \large
    \begin{tabular}{*{9}{c}}
        \toprule
       Layers locations & [3,6,9] & [2,5,9] &  [2,5,8] & [1,4,7] &  [1,6,9] \\
        \midrule
        Rank-1 & 76.7 & 76.2 {\color{red} (-0.5)} & 76.3 {\color{red} (-0.4)} & 75.3 {\color{red} (-1.4)} & 75.2 {\color{red} (-1.5)} \\
        mAP & 72.8 & 72.3 {\color{red} (-0.5)} & 72.5 {\color{red} (-0.3)} & 71.6 {\color{red} (-1.2)} & 71.4 {\color{red} (-1.4)} \\
        \bottomrule
    \end{tabular}
    }
\caption{Analysis of token sparsification layers on Occluded-Duke dataset.}
\label{layers}
\end{table}

\subsection{More Implementation Details}
We adopted a two-phase training strategy to optimize computational memory efficiency. Initially, we focused on training the sparse encoder. Upon completion of this phase, the parameters of the sparse encoder were fixed, and we proceeded to train the parameters of the transformer decoder module. 
The first stage effectively circumvents memory constraints associated with large gallery memory storage. Subsequently, the second stage demonstrated a reduced need for epochs to achieve effective convergence.
Furthermore, we input only the tokens of gallery neighbors into the transformer decoder for model inference.

\section{Further Analysis on Applicability} 
\subsection{Analysis for Real-World Applications} 
We recognize that there are many challenges to overcome before our approach can be applied to real-world Re-ID scenarios. It is important to note that these challenges are not unique to our approach, but rather common issues faced by most existing person Re-ID methods. 
These challenge are mainly:
1) large number of gallery images; 2) the gallery needs update; 
3) expensive cost of feature storage; 4) no positive samples for a query image.

Classic person Re-ID techniques, such as pair-wise similarity calculation~\cite{ye2021deep} or gallery feature reconstruction/aggregation~\cite{xu2022learning, yu2021neighbourhood, liao2021transmatcher, wu2022learning}, also face the same issues of large gallery size, gallery update requirement, high cost of feature storage, and absence of positive samples in the gallery for a query image.
Moreover, for 1) and 3), in our implementation, the feature size is 243x768 compared to the gallery image size 224x224x3, the extra feature storage is still in the same order of magnitude. For 2), our approach treats each image in the gallery as independent during matching and consolidation, so an update of the gallery would just simply involve deleting and inserting operations. For 4), If there are no positive samples in the gallery for a query image, it is impossible for person Re-ID approaches, including ours, to find the correct identity, as it would violate the pre-conditions for the task. In such cases, it is likely that our approach would use the closest negative samples for consolidation and report a low confidence score.

In addition, it should be noted that our approach still has the potential to be applicable to real-world scenarios. To address the limitations outlined in points 1) and 2), we can employ the efficient HNSW algorithm~\cite{malkov2018efficient}, which has demonstrated fast search speeds of 94ms on a million-scale SIFT dataset and 18.3s on a million-scale ImageNet dataset. To tackle point 3), we can replace fine-grained patch tokens with a $\mathrm{[cls]}$ token, which greatly reduces memory storage requirements while still maintaining an acceptable level of accuracy, as in paper Sec. 4.3. Finally, in regards to point 4), we can adaptively adjust the size of neighbors based on EMD distance rather than relying on a fixed number of neighbors. Nevertheless, our framework is robust to the use of incorrect neighbor features, as demonstrated by our results in Fig. 3(b) of the paper, where an over-sized choice of neighbors (K=14) had only a minimal impact on performance (a decrease of 1.1\% in mAP and 2.1\% in Rank-1 compared to the best results).

\begin{table}[h]
\centering
    \resizebox{0.95\columnwidth}{!}{
    \large
    \begin{tabular}{*{5}{c}}
        \toprule
       Gallery Feature & Gallery Images & Memory Cost (G) & Rank-1 (\%) &  mAP (\%)  \\
        \midrule
        $\mathrm{[cls]}$ token & 17,661 & 0.05 & 75.9 & 70.6  \\
        patch token + $\mathrm{[cls]}$ token & 17,661 & 12.28 & 76.7 & 72.8 \\
        \bottomrule
    \end{tabular}
        }
\caption{The analysis of using [cls] token in feature consolidation to save storage.}
\label{gallery-feature}
\end{table}

\subsection{Time Analysis of Distance Computation}
As described in Sec. 3.4 of the paper, our method uses a two-step procedure to efficiently select optimal samples. The first step involves finding 100 candidate samples using cosine distance and the second step integrates EMD to determine the final optimal samples. Our approach showed its efficiency on the Occluded-Duke dataset, where computing EMD for all 17,661 gallery images took 1152.5s, while our two-stage process took only 11.0s, as presented in ~\cref{running-time}. This is in comparison to the k-reciprocal re-ranking process commonly used in Re-ID, which takes 157.1s, making our EMD computation highly efficient.

\begin{table}[h]
\centering
    \resizebox{0.9\columnwidth}{!}{
    \large
    \begin{tabular}{lccccc}
        \toprule
     & \multicolumn{3}{c}{Time Cost (s)}  & \multicolumn{2}{c}{Metric (\%)}  \\
       \multirow{-2}{*}{Method} & Cosine  & EMD  & Total  & Rank-1     & mAP  \\
        \midrule
        Bruteforce & 0.5 & 1152.5 & 1153.0 & 76.7 & 72.8 \\
        Two-Stage (Ours) & 0.5 & 10.5 & 11.0 & 76.8 & 72.7 \\
        \bottomrule
    \end{tabular}
    }
    \label{2-run-time}
\caption{Running time analysis for the distance computation.}
\label{running-time}
\end{table}

\section{Future Work}
\subsection{Optimization for Drop Rate}
 We acknowledge that the patch drop rate in our model is currently fixed as a potential limitation. Our paper ablation studies on the keep ratio have shown that a range of 0.7-0.9 consistently yields superior results on Occluded-Duke. While not currently implemented, future research plans include the adaptive optimization of the drop rate.

\subsection{Generalizability of Framework}
In the future, we envisage investigating the proposed pruning and consolidation framework across a wider spectrum of occlusion challenges, including but not limited to mixed holistic and occluded ReID, occluded object classification, among others. This endeavor is aimed at bolstering the generalizability and applicability of our framework.

\newpage

\end{document}